\documentclass[lettersize,journal]{IEEEtran}
\usepackage{amsmath,amsfonts}
\usepackage{algorithmic}
\usepackage{algorithm}
\usepackage{array}
\usepackage[caption=false,font=normalsize,labelfont=sf,textfont=sf]{subfig}
\usepackage{textcomp}
\usepackage{stfloats}
\usepackage{url}
\usepackage{verbatim}
\usepackage{graphicx}
\usepackage{cite}

\usepackage{color}
\usepackage{multirow}
\usepackage{pifont}
\usepackage[pagebackref,breaklinks,colorlinks]{hyperref}

\begin{document}

\title{STDAN: Deformable Attention Network for Space-Time Video Super-Resolution}

\author{Hai~Wang, Xiaoyu Xiang, Yapeng Tian, Wenming Yang, and Qingmin Liao
\thanks{H.~Wang, W.~Yang and Q.~Liao are with Tsinghua Shenzhen International Graduate School / Department of Electronic Engineering, Tsinghua University, Shenzhen 518055, China (e-mail: hwangshenzhen@163.com; yangelwm@163.com; liaoqm@tsinghua.edu.cn).}
\thanks{X.~Xiang is with the On-Device AI team at Meta Reality Labs, Menlo Park, CA 94025 USA (e-mail: xiangxiaoyu@fb.com).}
\thanks{Y.~Tian is with the Department of Computer Science, University of Rochester, Rochester, NY 14627 USA (e-mail: yapengtian@rochester.edu).}
}


\markboth{Journal of \LaTeX\ Class Files,~Vol.~14, No.~8, August~2021}%
{Shell \MakeLowercase{\textit{et al.}}: A Sample Article Using IEEEtran.cls for IEEE Journals}


\maketitle

\begin{abstract}
The target of space-time video super-resolution (STVSR) is to increase the spatial-temporal resolution of low-resolution (LR) and low frame rate (LFR) videos. Recent approaches based on deep learning have made significant improvements, but most of them only use two adjacent frames, that is, short-term features, to synthesize the missing frame embedding, which cannot fully explore the information flow of consecutive input LR frames. In addition, existing STVSR models hardly exploit the temporal contexts explicitly to assist high-resolution (HR) frame reconstruction. To address these issues, in this paper, we propose a deformable attention network called STDAN for STVSR. First, we devise a long-short term feature interpolation (LSTFI) module, which is capable of excavating abundant content from more neighboring input frames for the interpolation process through a bidirectional RNN structure. Second, we put forward a spatial-temporal deformable feature aggregation (STDFA) module, in which spatial and temporal contexts in dynamic video frames are adaptively captured and aggregated to enhance SR reconstruction. Experimental results on several datasets demonstrate that our approach outperforms state-of-the-art STVSR methods. The code is available at \href{https://github.com/littlewhitesea/STDAN}{https://github.com/littlewhitesea/STDAN}.
\end{abstract}


\section{Introduction}
\IEEEPARstart{T}{he} goal of space-time video super-resolution (STVSR) is to reconstruct photo-realistic high-resolution (HR) and high frame rate (HFR) videos from corresponding low-resolution (LR) and low frame rate (LFR) ones. STVSR methods have attracted much attention in the computer vision community since HR slow-motion videos provide more visually appealing content for viewers. Many traditional algorithms \cite{sp-stvsr,2005-stvsr,2010-stvsr1,2010-stvsr2,2011-stvsr} are proposed to solve the STVSR task. However, due to their strict assumptions in their manually designed regularization, these methods mostly suffer from ubiquitous object and camera motions in videos.

In recent years, deep learning approaches have made great progress in diverse low-level visual tasks \cite{tdan,zhao2022vfi,dain,sepconv,zhang2020sisr,fastdvdnet,vdeblur,zheng2020derain}. Particularly, video super-resolution (VSR) \cite{tdan,edvr} and video frame interpolation (VFI) \cite{dain,superslomo} networks among these approaches can be combined together to tackle STVSR. Specifically, the VFI model interpolates the missing LR video frames. Then, the VSR model can be adopted to reconstruct HR frames. Nevertheless, the two-stage STVSR approaches usually have large model sizes, and the essential association between the temporal interpolation and spatial super-resolution is not explored.

\begin{figure}[t]
\centering
\includegraphics[width=\linewidth]{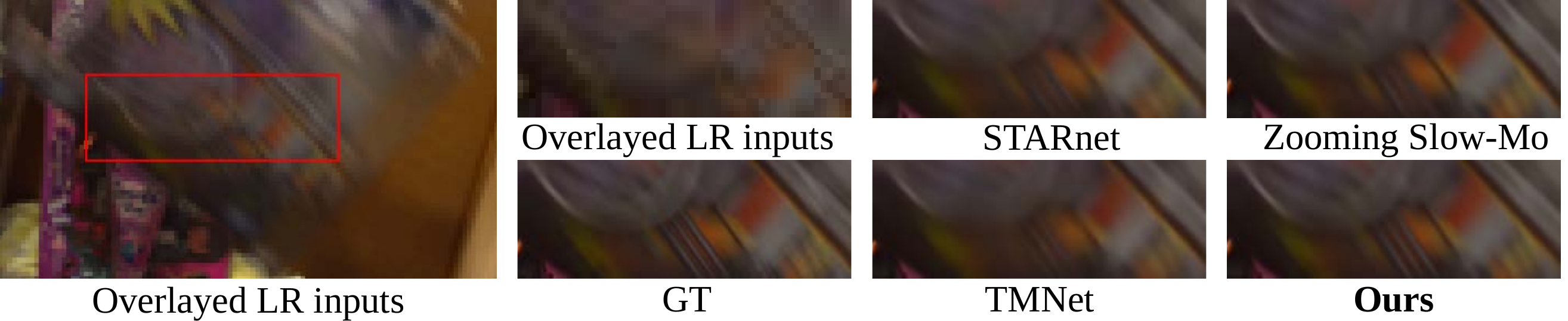}
\caption{Example of space-time video super-resolution (STVSR). Compared with three recent SOTA STVSR methods, our network can reconstruct more accurate structures.}
\label{fig:first}
\end{figure}

To build an efficient model and explore mutual information between temporal interpolation and spatial super-resolution, several one-stage STVSR networks \cite{zooming,tmnet,stvsr-eccv20,geng2022rstt} are proposed. These approaches can simultaneously handle the space and time super-resolution of videos in diverse scenes. Most of them only leverage corresponding two adjacent frames for interpolating the missing frame feature. However, other neighboring input LR frames can also contribute to the interpolation process. In addition, existing one-stage STVSR networks are limited in fully exploiting spatial and temporal contexts among various frames for SR reconstruction. To alleviate these problems, in this paper, we propose a one-stage framework named STDAN for STVSR, which is superior to recent methods, illustrated in Fig. \ref{fig:first}. The cores of STDAN are (1) a feature interpolation module known as Long-Short Term Feature Interpolation (LSTFI), and (2) a feature aggregation module known as Spatial-Temporal Deformable Feature Aggregation (STDFA).

The LSTFI module, composed of long-short term cells (LSTCs), utilizes a bidirectional RNN \cite{bi-rnn} structure to synthesize features for missing intermediate frames. Specifically, to interpolate the intermediate feature, we adopt the forward and backward deformable alignment \cite{zooming} for dynamically sampling two neighboring frame features. Then, the preliminary intermediate feature in the current LSTC is mingled with the hidden state that contains long-term temporal context from previous LSTCs to obtain the final interpolated features. 

The STDFA module aims to capture spatial-temporal contexts among different frames to enhance SR reconstruction. To dynamically aggregate the spatial-temporal information, we propose to use deformable attention to adaptively discover and leverage relevant spatial and temporal information. The process of STDFA can be divided into two phases: cross-frame spatial aggregation and adaptive temporal aggregation. Through deformable attention, the cross-frame spatial aggregation phase dynamically fuses useful content from different frames. The adaptive temporal aggregation phase mixes the temporal contexts among these fused frame features further to acquire enhanced features.

The contributions of this work are three-fold: (1) We design a deformable attention network (STDAN) to deal with STVSR. Our STDAN with fewer parameters achieves state-of-the-art performance on multiple datasets; (2) We propose a long-short term feature interpolation module, where abundant information from more neighboring frames are explored for the interpolation process of missing frame features; (3) We put forward a spatial-temporal deformable feature aggregation module, which can dynamically capture spatial and temporal contexts among video frames for enhancing features to reconstruct HR frames.

\section{Related Work}

In this section, we discuss some relevant works on video super-resolution, video frame interpolation, and space-time video super-resolution.

\noindent
\textbf{Video Super-Resolution.}
The goal of video super-resolution (VSR) \cite{tga,edvr,vsrnet,basicvsr} is to generate temporally coherent high-resolution (HR) videos from corresponding low-resolution (LR) ones. Since input LR video frames are consecutive, many researchers focus on how to aggregate the temporal contexts from the neighboring frames for super-resolving the reference frame. Several VSR approaches \cite{toflow,vespcn,drvsr,sofvsr,rbpn} adopt optical flow to align the reference frame with neighboring video frames. Nevertheless, the estimated optical flow may be inaccurate due to the occlusion and fast motions, leading to poor reconstruction results. To avoid using optical flow, deformable convolution \cite{dcnv1,dcnv2} is applied in \cite{tdan,d3dnet,edvr} to perform the temporal alignment in a feature space. In addition, Li \emph{et al.} \cite{mucan} established a multi-correspondence aggregation network to exploit similar patches between and within frames. Dynamic filters \cite{duf} and non-local \cite{pfnl,dnln} modules are also exploited to aggregate the temporal information. 

\noindent
\textbf{Video Frame Interpolation.}
Video frame interpolation (VFI) \cite{memcnet,adacof,bmbc,dain,zhao2022vfi} aims to synthesize the missing intermediate frame with two adjacent video frames, which is extensively used in slow-motion video generation. Specifically, for generating the intermediate frame, U-Net structure modules \cite{superslomo} are employed to compute optical flows and visibility maps between two input frames. To cope with occlusion in VFI, contextual features \cite{ctxsyn} are further introduced into the interpolation process. Furthermore, Bao \emph{et al.} \cite{dain} proposed a depth-aware module to detect occlusions explicitly for VFI. On the other hand, unlike most VFI methods using optical flow, Niklaus \emph{et al.} \cite{adaconv,sepconv} adopted the adaptive convolution to predict kernels directly and then leveraged these kernels to estimate pixels of the intermediate video frame. Recently, attention mechanism \cite{dain} and deformable convolution \cite{dcn-vfi,adacof} are explored.

\noindent
\textbf{Space-Time Video Super-Resolution.}
Compared to video super-resolution, space-time video super-resolution (STVSR) needs to implement super-resolution in time and space dimensions. Due to strict assumptions and manual regularization, conventional STVSR methods \cite{2005-stvsr,2010-stvsr1,2011-stvsr} cannot effectively process the spatial-temporal super-resolution of sophisticated LR input videos. In recent years, significant advances have been made from the deep neural network (DNN). Through merging VSR and VFI into a joint framework, Kang \emph{et al.} \cite{stvsr-eccv20} put forward a DNN model for STVSR. To exploit mutually informative relationships between time and space dimensions, STARnet \cite{starnet} with an extra optical flow branch is proposed to generate HR slow-motion videos. In addition, Xiang \emph{et al.} \cite{zooming} developed a deformable ConvLSTM \cite{convlstm} module, which can achieve sequence-to-sequence (S2S) learning in STVSR. Base on \cite{zooming}, Xu \emph{et al.} \cite{tmnet} proposed a temporal modulation block to perform controllable STVSR. Recently, Geng \emph{et al.} \cite{geng2022rstt} proposed a STVSR network based on Swin Transformer. However, most of them only leverage two adjacent frame features to interpolate the intermediate frame feature, and they hardly explore spatial and temporal contexts explicitly among video frames. To address these problems, we propose a spatial-temporal deformable network (1) to use more content from input LR frames for the interpolation process and (2) employ deformable attention to dynamically capture spatial-temporal contexts for HR frame reconstruction.

\begin{figure*}[t]
\centering
\includegraphics[height=4.4cm]{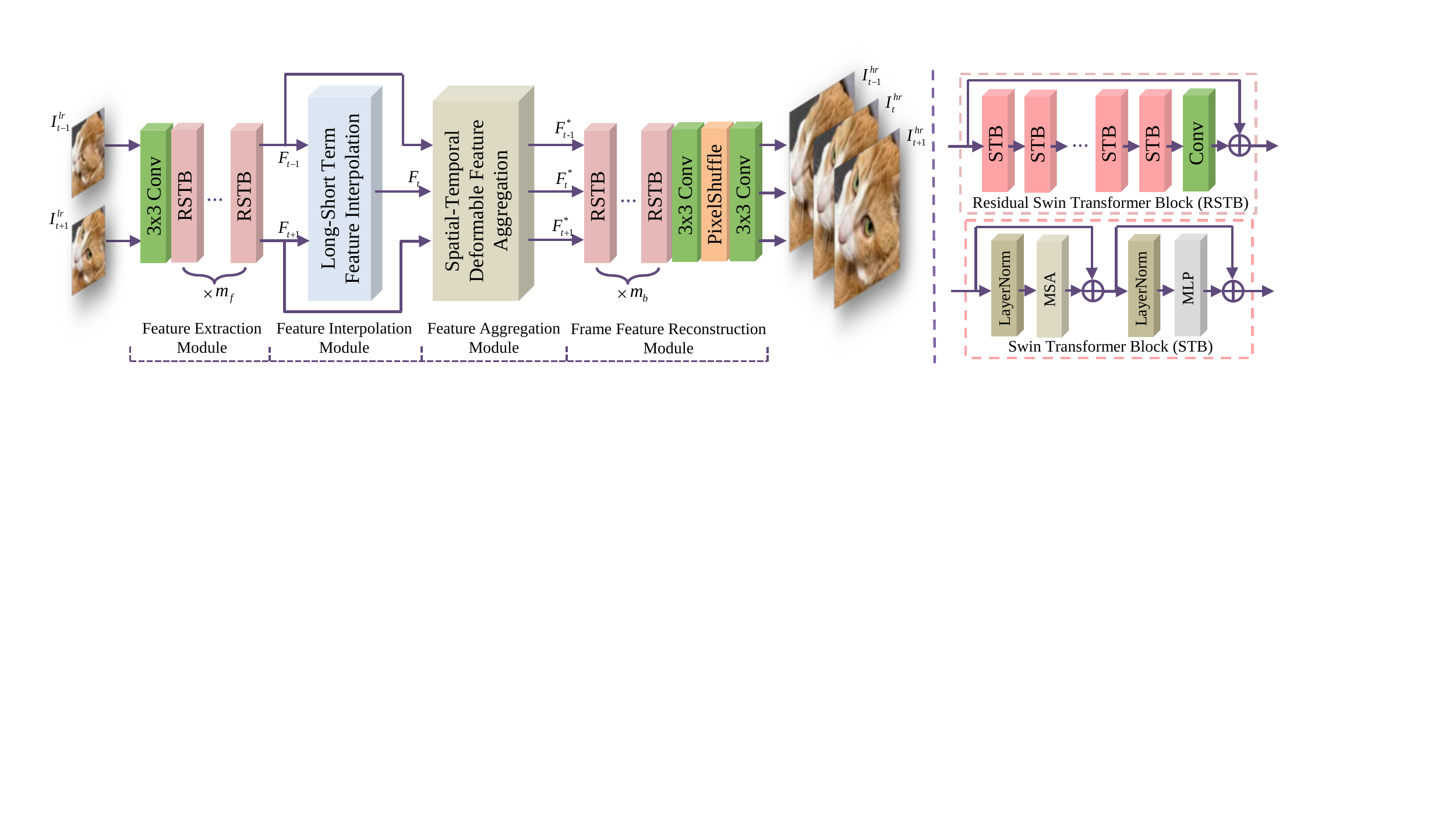}
\caption{The architecture of our proposed STDAN. Long-short term feature interpolation is capable of exploring more neighboring LR frames to synthesize the intermediate frame in the feature space. Spatial-temporal deformable feature aggregation is utilized to capture spatial-temporal contexts by deformable attention. This figure only shows two input LR video frames from a long video sequence for a presentation.}
\label{fig:whole_net}
\end{figure*}

\section{Our Method}

The architecture of our proposed network is illustrated in Fig. \ref{fig:whole_net}, which consists of four parts: feature extraction module, long-short term feature interpolation (LSTFI) module, spatial-temporal deformable feature aggregation (STDFA) module and frame feature reconstruction module. Given a low-resolution (LR) and low frame rate (LFR) video with $N$ frames: $\left\{I_{2t-1}^{lr}\right\}_{t=1}^{N}$, our STDAN can generate $2N-1$ consecutive high-resolution (HR) and high frame rate (HFR) frames: $\left\{I_{t}^{hr}\right\}_{t=1}^{2N-1}$. The structure of each module is described in the following.

\subsection{Frame Feature Extraction}

We first use a $3\times3$ convolutional layer in the feature extraction module to get shallow features $\left\{F_{2t-1}^{s}\right\}_{t=1}^{N}$ from the $N$ input LR video frames. Considering that these shallow features lack long-range spatial information due to the locality of the naive convolutional layer, which may cause poor quality in the next feature interpolation module. We hope to extract these shallow features further to establish the correlation between two distant locations.

Recently, Transformer-based models have realized good performance in computer vision \cite{detr,autoformer,vit,tracking} owing to the strong capacity of Transformer to model long-range dependency. However, the computation cost of self-attention in Transformer is high, which limits its extensive application in video-related tasks. To overcome the drawback, Liu \emph{et al.} \cite{swin} put forward Swin Transformer block (STB) to achieve linear computational complexity with respect to image size. Based on efficient and effective STB \cite{swin}, Liang \emph{et al.} \cite{swinir} proposed residual Swin Transformer block (RSTB) to construct SwinIR for image restoration. Thanks to the powerful ability to model long-range dependency of RSTB, SwinIR \cite{swinir} obtains state-of-the-art (SOTA) performance compared with CNN-based methods. In this paper, to acquire features $\left\{F_{2t-1}\right\}_{t=1}^{N}$ that capture long-range spatial information, we also use $m_f$ RSTBs \cite{swinir} to extract shallow features $\left\{F_{2t-1}^{s}\right\}_{t=1}^{N}$ further, illustrated in Fig. \ref{fig:whole_net}. We can see that the RSTB is a residual block with several STBs and one convolutional layer. In addition, given a tensor $X_{in}$ as input, the detailed process of STB to output $X_{out}$ is formulated as:
\begin{equation}\label{eq:concate}
\begin{split}
    & X = MSA(LayerNorm(X_{in})) + X_{in}, \\
    & X_{out} = MLP(LayerNorm(X)) + X,
\end{split}
\end{equation}
where \emph{MSA} and $X$ denotes multi-head self-attention module and intermediate results, respectively.

\begin{figure}
\centering
\includegraphics[height=5cm]{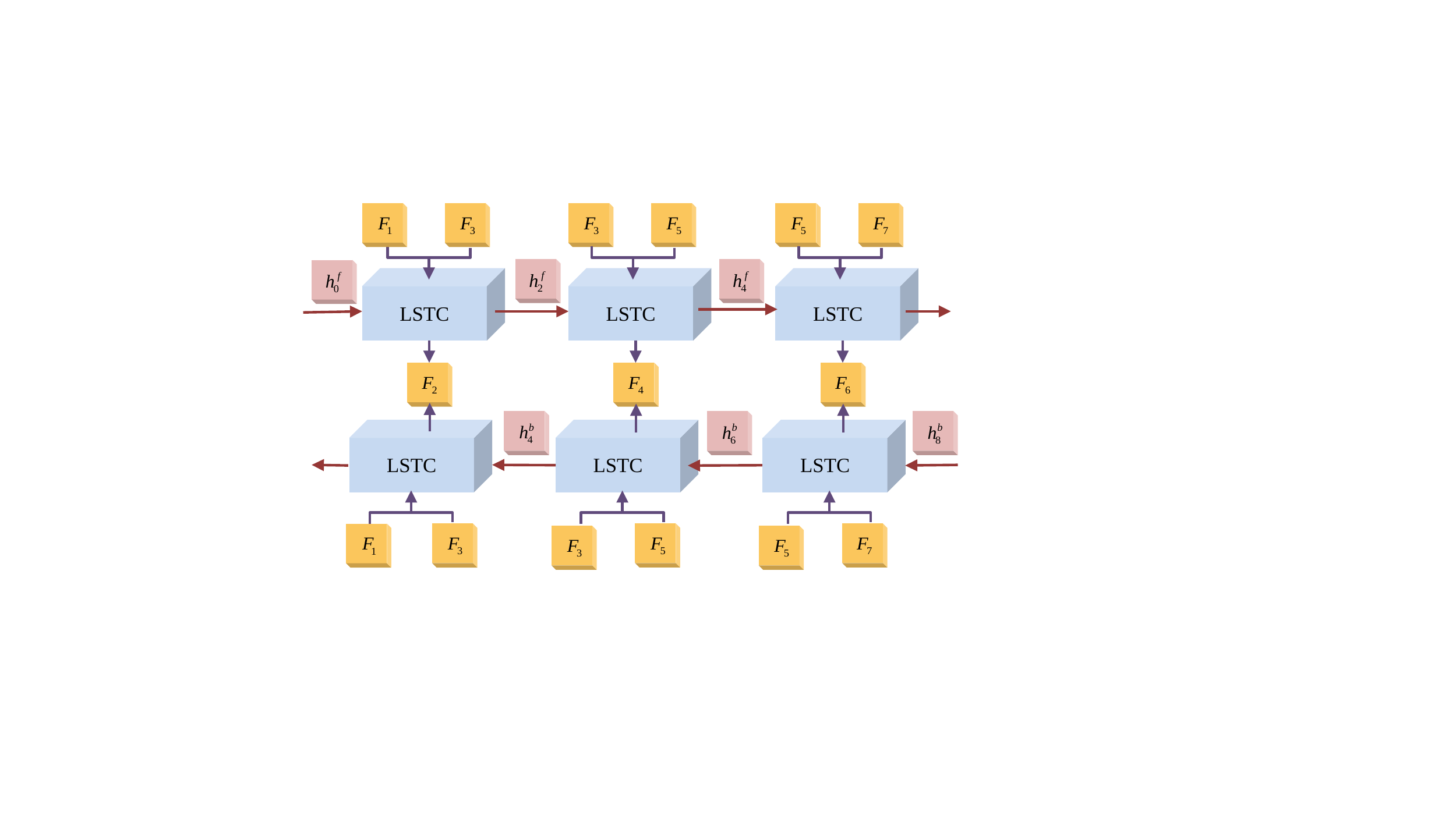}
\caption{The framework of our long-short term feature interpolation (LSTFI) module. It consists of long-short term cells (LSTCs) with bidirectional RNN, which can fully exploit the whole input video frame features during the interpolation process. Note that the two neighboring frame features and the hidden state from previous LSTC provide short-term and long-term content for interpolation results, respectively. Here, $h^f_0$ and $h^b_8$ denote initialized hidden states for the forward and backward recurrent propagation, respectively. More specifically, the $h^f_0$ serves as the forward hidden state for predicting the first missing frame feature: $F_2$, while the $h^b_8$ is regarded as the backward hidden state for predicting the last missing frame feature: $F_6$.}
\label{fig:lstfi}
\end{figure}

\subsection{Long-Short Term Feature Interpolation}

To implement the super-resolution in the time dimension, we also utilize a feature interpolation module to synthesize the intermediate frames in the LR feature space, like \cite{zooming,tmnet}. Specifically, given the two extracted features: $F_{1}$ and $F_{3}$, the feature interpolation module can synthesize the feature $F_{2}$ corresponding to the missing frame $I_{2}^{lr}$. Generally, to obtain the intermediate feature, we should capture the pixel-wise motion information first. Optical flow is usually adopted to estimate the motion between video frames. However, there are several shortcomings in using optical flow for interpolation. The computational cost is high to calculate optical flow precisely, and estimated optical flow may be inaccurate due to the occlusion or motion blur, which causes poor interpolation results.

Considering the drawback of optical flow, Xiang \emph{et al.} \cite{zooming} employed multi-level deformable convolution \cite{edvr} to perform frame feature interpolation. The learned offset used in deformable convolution can implicitly capture forward and backward motion information and achieve good performance. However, the synthesis of intermediate frame feature \cite{zooming,tmnet} only utilizes the two neighboring frame features, which cannot fully explore the information from the other input frames to assist in the process. Unlike feature interpolation in previous STVSR algorithms \cite{zooming,tmnet}, we propose a long-short term feature interpolation (LSTFI) module to realize the intermediate frame in our STDAN, which is capable of exploiting helpful information from more input frames.

\begin{figure*}[t]
\centering
\includegraphics[height=6cm]{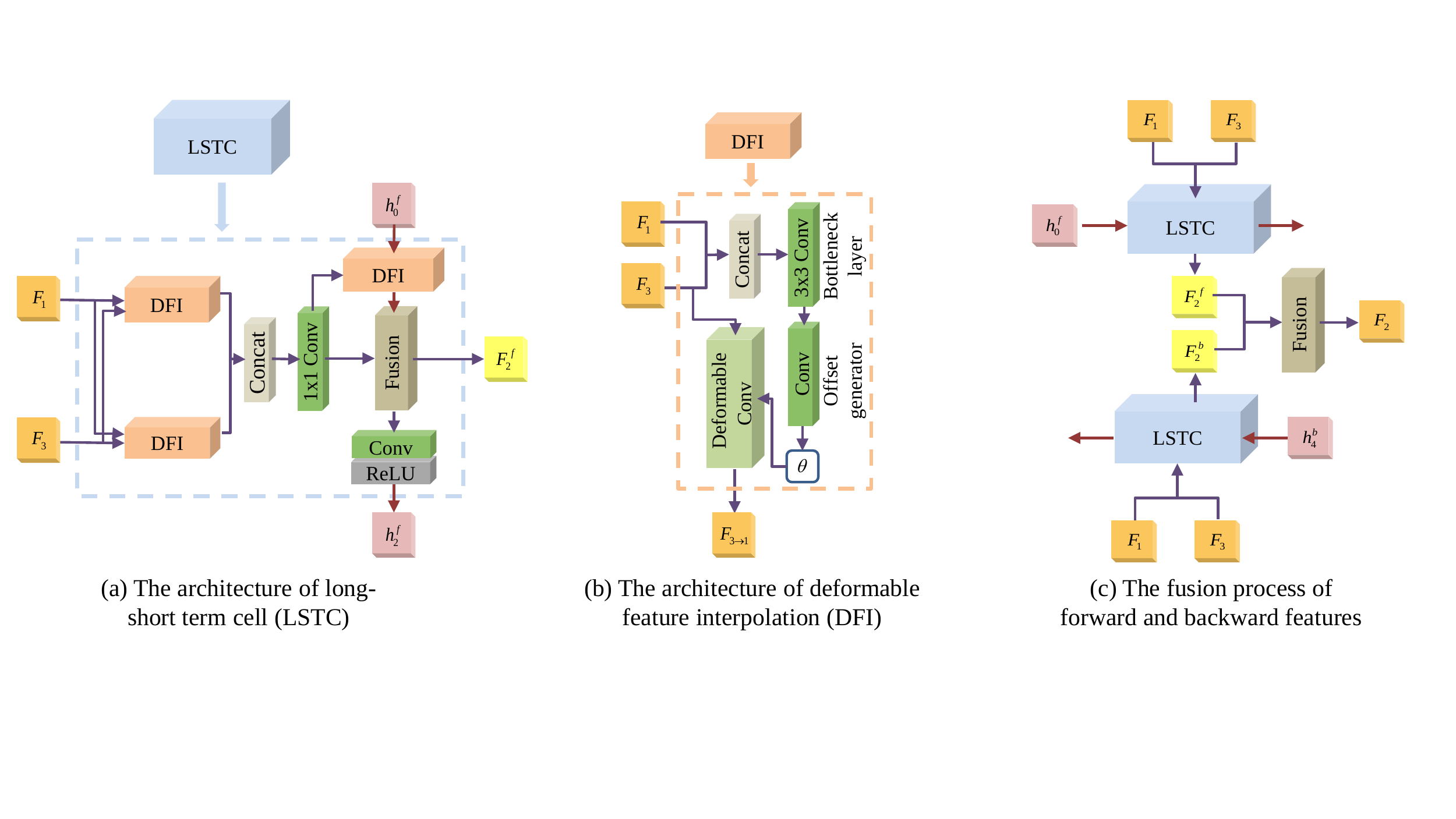}
\caption{Overview of the proposed LSTC and fusion process of interpolation results from forward and backward branches. We adopt DFI block \cite{zooming} to adaptively align the hidden state from the previous LSTC with the current preliminary interpolation result. Note that the final intermediate frame feature is achieved by fusing the interpolation results from forward and backward branches.}
\label{fig:lstcell}
\end{figure*}

As illustrated in Fig. \ref{fig:lstfi}, we adopt a bidirectional recurrent neural network (RNN) \cite{bi-rnn} to construct the LSTFI module, which consists of two branches in forward and backward directions. Take the forward branch as an example. Two neighboring frame features and the hidden state from the previous long-short term cell (LSTC) are fed into each LSTC, and then the LSTC generates the corresponding intermediate frame feature and current hidden state used for subsequent LSTC. Here, the two neighboring frame features and hidden state serve as short-term and long-term information for the intermediate feature, respectively. However, each branch's hidden state only considers the unidirectional information flow. To fully mine the information flow of these frame features for the interpolation procedure, we fuse interpolation results from LSTCs in the forward and backward branches to acquire the final intermediate frame feature.

The architecture of LSTC and the fusion process are shown in Fig. \ref{fig:lstcell}. Given two neighboring frame features $F_1$ and $F_3$, we employ deformable feature interpolation (DFI) block \cite{zooming} to capture the forward and backward motion between the two features implicitly. For simplification, we take the feature $F_{3\rightarrow{1}}$ that has experienced backward motion compensation as an example. As illustrated in Fig. \ref{fig:lstcell}(b), the two frame features are concatenated along channel dimension, and then pass through offset generation function $H_{og}^{b}$ to predict an offset with backward motion information:
\begin{equation}\label{eq:offset_back}
    \theta_{3\rightarrow{1}} = H_{og}^{b}([F_3,F_1])\ ,
\end{equation}
where $H_{og}^{b}$ consists of convolutional layers, and $[.,.]$ denotes the concatenation along channel dimension. With the learned offset, we adopt deformable convolution \cite{dcnv2} as motion compensation function to obtain compensated feature:
\begin{equation}\label{eq:dcn_back}
    F_{3\rightarrow{1}} = DConv([F_3,\theta_{3\rightarrow{1}}])\ ,
\end{equation}
where $DConv$ denotes the operation of deformable convolution.

To blend the features $F_{1\rightarrow{3}}$ and $F_{3\rightarrow{1}}$ that have experienced forward and backward motion compensation respectively, a $1\times{1}$ convolutional layer is applied, which can perform pixel-level linear weighting to achieve preliminary interpolation feature $F_2^{p}$. Note that the acquisition of feature $F_2^{p}$ only utilizes the short-term information. In order to combine long-term information $h_0^{f}$, the hidden state from the previous LSTC, we first use the other DFI block to align $h_0^{f}$ with the current feature $F_2^{p}$, since there may be some misalignment. The process is expressed as:
\begin{equation}\label{eq:hidden_align}
    h_{0\rightarrow{2}}^f = DAlign(h_0^{f},F_2^{p})\ ,
\end{equation}
where $DAlign(.)$ indicates the operation of DFI block. At the end of LSTC, we apply a fusion function into aligned hidden state $h_{0\rightarrow{2}}^f$ and preliminary interpolation result $F_2^{p}$ to obtain forward intermediate feature:
\begin{equation}\label{eq:forward_fea}
    F_2^{f} = H_{fs}(F_2^{p},h_{0\rightarrow{2}}^f)\ ,
\end{equation}
where $H_{fs}$ refers to the fusion function. Then, the intermediate feature $F_2^{f}$ passes through a convolutional layer and an activation layer in sequence to produce hidden state $h_2^{f}$ for the subsequent LSTC.

For fully exploring the whole input frame features for interpolation, the bidirectional RNN structure is utilized in our LSTFI module, so we fuse the forward intermediate feature $F_2^{f}$ and backward intermediate feature $F_2^{b}$ to get the final intermediate frame feature $F_2$, shown in Fig. \ref{fig:lstcell}(c).

\subsection{Spatial-Temporal Deformable Feature Aggregation}

With the assistance of the LSTFI module, we now have $2N-1$ frame features, where the generation of $N-1$ intermediate frame features combines their adjacent frame features with hidden states. Although the hidden states can introduce certain temporal information, the whole interpolation procedure hardly explicitly explores the temporal information between various frames. In addition, the $N$ input frame features are merely processed independently in the feature extraction module. However, these frame features $\left\{F_{t}\right\}_{t=1}^{2N-1}$ are consecutive, which means there are abundant temporal content without being exploited among these features. 

For a feature vector $\mathbf{f}_t$ whose location is $\mathbf{p}_o$ on feature $F_{t}$, the simplest method to aggregate temporal information is adaptive fusion with the feature vector on the same location from the other $2N-2$ frame features. However, the aggregation approach has several drawbacks. Generally, the corresponding point on other frame features may not be on the same location due to inter-frame motion. Furthermore, there are multiple helpful feature vectors for $\mathbf{f}_t$ from each of the $2N-2$ frame features. Based on the above analysis, we propose a spatial-temporal deformable feature aggregation (STDFA) module to mix cross-frame spatial information adaptively and capture the long-range temporal information.

\begin{figure*}
\centering
\includegraphics[height=4.4cm]{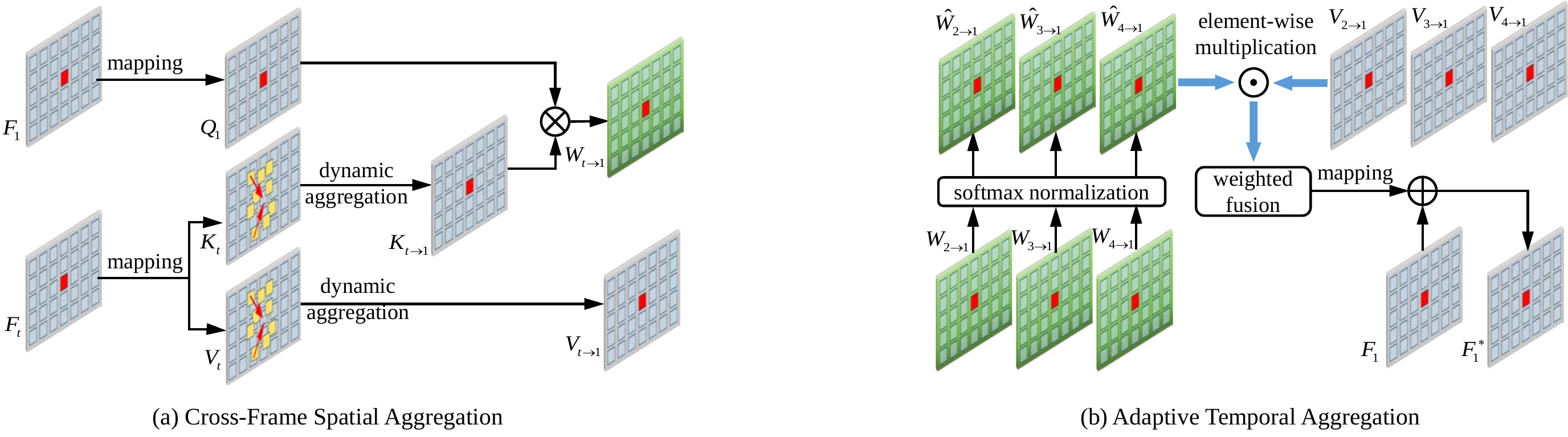}
\caption{The detailed process of spatial-temporal deformable feature aggregation (STDFA) module. Note that we only show the case when the number of frame features is 4. Under the case, the value of $t$ can be 2, 3 or 4 for frame feature $F_1$.}
\label{fig:stdfa}
\end{figure*}

Specifically, we utilize the STDFA module to learn the residual auxiliary information from the remaining $2N-2$ frame features for each frame feature $F_{t}$. As presented in Fig. \ref{fig:stdfa}, the processing of the STDFA module can be divided into two parts: spatial aggregation and temporal aggregation. To adaptively fuse cross-frame spatial content of frame feature $F_i$ from the other frame features, we perform deformable attention to each pair: $F_i$ and $F_j$ ($j\in[1, 2N-1]$, $j\neq{i}$). In detail, frame feature $F_i$ passes through a linear layer to get embedded feature $Q_i$. Similarly, frame feature $F_j$ is fed into two linear layers to obtain embedded features $K_j$ and $V_j$, respectively. 

To implement deformable attention between $F_i$ and $F_j$, we first predict the offset map:
\begin{equation}\label{eq:offset_generation}
    \Delta{M_{j\rightarrow{i}}} = H_{og}([Q_i,K_j])\ ,
\end{equation}
where $H_{og}$ indicates offset generation function consisting of several convolutional layers with $k\times{k}$ kernel. The offset map $\Delta{M_{j\rightarrow{i}}}$ at position $\mathbf{p}_o$ is expressed as:
\begin{equation}\label{eq:offset_point}
    \Delta{M_{j\rightarrow{i}}(\mathbf{p}_o)} = [\Delta{\mathbf{p}_1}, \Delta{\mathbf{p}_2}, \cdots, \Delta{\mathbf{p}_{\xi}}, \cdots, \Delta{\mathbf{p}_{k^2}} ]\ .
\end{equation}
Then the offsets $\Delta{M_{j\rightarrow{i}}(\mathbf{p}_o)}$ are combined with $k^2$ pre-specified sampling locations to perform deformable sampling. Here, we denote the pre-specified sampling location as $\mathbf{p}_{\xi}$, and the value set of $\mathbf{p}_{\xi}$ of $k\times{k}$ kernel is defined as: 
\begin{equation}\label{eq:offset_range}
\mathbf{p}_{\xi}\in{\left\{(-{\left \lfloor \frac{k}{2} \right \rfloor},-{\left \lfloor \frac{k}{2} \right \rfloor}), \cdots, ({\left \lfloor \frac{k}{2} \right \rfloor},{\left \lfloor \frac{k}{2} \right \rfloor}) \right\}} \ ,
\end{equation}
where $\left \lfloor \cdot \right \rfloor$ denotes rounding down function. 

With the offsets $\Delta{M_{j\rightarrow{i}}}(\mathbf{p}_o)$,  the embedded feature vector $Q_i(\mathbf{p}_o)$ can attend $k^2$ related points in $K_j$. Nevertheless, not all the information of these $k^2$ points is helpful for $Q_i(\mathbf{p}_o)$. In addition, each point on embedded feature $Q_i$ needs to search $k^2$ points, which inevitably causes a large storage occupation. To avoid irrelevant points and reduce storage occupation, we only choose the first $T$ points that are most relevant. To select the $T$ points, we calculate the inner product between two embedded feature vectors as the relevance score:
\begin{equation}\label{eq:relevance_score}
    RS_{j\rightarrow{i}}(\mathbf{p}_o,\xi) =  Q_i(\mathbf{p}_o) \cdot K_j(\mathbf{p}_o + \mathbf{p}_{\xi} + \Delta{\mathbf{p}_{\xi}})\ ,
\end{equation}
The larger the score, the more relevant the two points are. According to this criterion, we can determine the $T$ points. In the following, to distinguish the selected $T$ points from original $k^2$ points, we denote the pre-specified sampling location and learned offset of the $T$ points as $\mathbf{\overline{p}}_{\xi}$ and $\Delta{\mathbf{\overline{p}}_{\xi}}$, respectively.

To adaptively mingle the spatial information from the $T$ locations for each embedded feature vector $Q_i(\mathbf{p}_o)$, we first adopt softmax function to calculate the weight of these points:
\begin{equation}\label{eq:weight_point}
    w_{\xi} = \frac{e^{Q_i(\mathbf{p}_o) \cdot K_j(\mathbf{p}_o + \mathbf{\overline{p}}_{\xi} + \Delta{\mathbf{\overline{p}}_{\xi}})}}{\sum_{\xi=1}^T{e^{Q_i(\mathbf{p}_o) \cdot K_j(\mathbf{p}_o + \mathbf{\overline{p}}_{\xi} + \Delta{\mathbf{\overline{p}}_{\xi}})}}}\ . 
\end{equation}
Then, with the weights and the embedded feature vector $K_j(\mathbf{p}_o + \mathbf{\overline{p}}_{\xi} + \Delta{\mathbf{\overline{p}}_{\xi}})$, we can obtain corresponding updated embedded feature vector:

\begin{equation}\label{eq:update_key}
    K_{j\rightarrow{i}}(\mathbf{p}_o) = \sum_{\xi=1}^T{w_{\xi}\cdot{K_j(\mathbf{p}_o + \mathbf{\overline{p}}_{\xi} + \Delta{\mathbf{\overline{p}}_{\xi}})}}\ .
\end{equation}
Same as $K_{j\rightarrow{i}}(\mathbf{p}_o)$, the updated vector $V_{j\rightarrow{i}}(\mathbf{p}_o)$ can be also achieved with the weight $w_{\xi}$. Finally, we calculate the updated relevant weight map $W_{j\rightarrow{i}}$ at each position $\mathbf{p}_o$ between $Q_i$ and $K_{j\rightarrow{i}}$ for the following temporal aggregation:
\begin{equation}\label{eq:update_weight}
    W_{j\rightarrow{i}}(\mathbf{p}_o) =  Q_i(\mathbf{p}_o) \cdot K_{j\rightarrow{i}}(\mathbf{p}_o)\ .
\end{equation}

To capture the temporal contexts of frame feature vector $F_i(\mathbf{p}_o)$ from the remaining $2N-2$ features, we also utilize softmax function to adaptively aggregating feature vectors $V_{j\rightarrow{i}}(\mathbf{p}_o)$. Specifically, the normalized temporal weight of each vector $V_{j\rightarrow{i}}(\mathbf{p}_o)$ ($j\in[1, 2N-1]$, $j\neq{i}$) is expressed as:
\begin{equation}\label{eq:temporal_weight}
    \hat{W}_{j\rightarrow{i}}(\mathbf{p}_o) = 
    \frac{e^{W_{j\rightarrow{i}}(\mathbf{p}_o)}}{\sum_{j=1}^{2N-1,j\neq{i}}{e^{W_{j\rightarrow{i}}(\mathbf{p}_o)}}}\ .
\end{equation}
Then, through fusing embedded feature vector $V_{j\rightarrow{i}}(\mathbf{p}_o)$ ($j\in[1, 2N-1]$, $j\neq{i}$) with the corresponding normalized weight, we can attain the embedded feature $V_{i}^{*}$ that aggregates the spatial and temporal contexts from other $2N-2$ embedded features. The weighted fusion process is defined as:
\begin{equation}\label{eq:st_aggregation}
    V_{i}^{*}(\mathbf{p}_o) = \sum_{j=1}^{2N-1,j\neq{i}}{
    \hat{W}_{j\rightarrow{i}}(\mathbf{p}_o)\cdot{V_{j\rightarrow{i}}(\mathbf{p}_o)}}\ .
\end{equation}
In the tail of STDFA module, the embedded feature $V_{i}^{*}$ is sent into a linear layer to acquire the residual auxiliary feature $F_{i}^{res}$. Finally, we add frame feature $F_{i}$ and residual auxiliary feature $F_{i}^{res}$ to get the enhanced feature $F_{i}^{*}$ that aggregates spatial and temporal contexts from the other $2N-2$ frame features. 



\begin{table*}[t]
\centering
\caption{Quantitative comparisons of our STDAN and other SOTA methods for space-time video super-resolution. The best two results are highlighted in \textcolor{red}{red} and \textcolor{blue}{blue} colors. Note that we conduct a padding operation to the input LR frames before feeding them into the networks, so the results of comparative methods on Vid4 are different from the reported results in the original papers.}
\setlength{\tabcolsep}{2mm}
\begin{tabular}{cc|cccccccccccc}
\hline
VFI               & (V)SR            & \multicolumn{2}{c}{Vid4} & \multicolumn{2}{c}{SPMC-11} & \multicolumn{2}{c}{Vimeo-Slow} & \multicolumn{2}{c}{Vimeo-Medium} & \multicolumn{2}{c}{Vimeo-Fast} & Speed & Parameters \\
Method            & Method         & PSNR       & SSIM        & PSNR         & SSIM         & PSNR          & SSIM           & PSNR           & SSIM            & PSNR          & SSIM           & FPS & (Million)  \\ \hline
SuperSloMo        & Bicubic        & 22.84      & 0.5772      & 24.91        & 0.6874       & 28.37         & 0.8102         & 29.94          & 0.8477          & 31.88         & 0.8793         & - & 19.8       \\
SuperSloMo        & RCAN           & 23.78      & 0.6385      & 26.50        & 0.7527       & 30.69         & 0.8624         & 32.50          & 0.8884          & 34.52         & 0.9076         & 2.49 & 19.8+16.0  \\
SuperSloMo        & RBPN           & 24.00      & 0.6587      & 26.14        & 0.7582       & 30.48         & 0.8584         & 32.79          & 0.8930          & 34.73         & 0.9108         & 2.06 & 19.8+12.7  \\
SuperSloMo        & EDVR           & 24.22      & 0.6700      & 26.46        & 0.7689       & 30.99         & 0.8673         & 33.85          & 0.8967          & 35.05         & 0.9136         & 6.85 & 19.8+20.7  \\ \hline
SepConv            & Bicubic        & 23.51      & 0.6273      & 25.67        & 0.7261       & 29.04         & 0.8290         & 30.61          & 0.8633          & 32.27         & 0.8890         & - & 21.7       \\
SepConv            & RCAN           & 24.99      & 0.7259      & 28.16        & 0.8226       & 32.13         & 0.8967         & 33.59          & 0.9125          & 34.97         & 0.9195         & 2.42 & 21.7+16.0  \\
SepConv            & RBPN           & 25.75      & 0.7829      & 28.65        & 0.8614       & 32.77         & 0.9090         & 34.09          & 0.9229          & 35.07         & 0.9238         & 2.01 & 21.7+12.7  \\
SepConv            & EDVR           & 25.89      & 0.7876      & 28.86        & \textcolor{blue}{0.8665}       & 32.96         & 0.9112         & 34.22          & 0.9240          & 35.23         & 0.9252         & 6.36 & 21.7+20.7  \\ \hline
DAIN              & Bicubic        & 23.55      & 0.6268      & 25.68        & 0.7263       & 29.06         & 0.8289         & 30.67          & 0.8636          & 32.41         & 0.8910         & - & 24.0       \\
DAIN              & RCAN           & 25.03      & 0.7261      & 28.15        & 0.8224       & 32.26         & 0.8974         & 33.82          & 0.9146          & 35.27         & 0.9242         & 2.23 & 24.0+16.0  \\
DAIN              & RBPN           & 25.76      & 0.7783      & 28.57        & 0.8598       & 32.92         & 0.9097         & 34.45          & 0.9262          & 35.55         & 0.9300         & 1.88 & 24.0+12.7  \\
DAIN              & EDVR           & 25.90      & 0.7830      & 28.77        & 0.8649       & 33.11         & 0.9119         & 34.66          & 0.9281          & 35.81         & 0.9323         & 5.20 & 24.0+20.7  \\ \hline
\multicolumn{2}{c|}{STARnet}       & 25.99      & 0.7819      & \textcolor{red}{29.04}        & 0.8509       & 33.10         & \textcolor{blue}{0.9164}         & 34.86          & 0.9356          & 36.19         & 0.9368         & 14.08 & 111.61      \\
\multicolumn{2}{c|}{Zooming Slow-Mo} & 26.14      & 0.7974      & 28.80        & 0.8635       & 33.36         & 0.9138         & 35.41          & 0.9361          & 36.81         & 0.9415         & \textcolor{red}{16.50} & 11.10       \\
\multicolumn{2}{c|}{RSTT}         & 26.20      & 0.7991      & 28.86        & 0.8634       & 33.50         & 0.9147         & \textcolor{blue}{35.66}          & \textcolor{blue}{0.9381}          & 36.80         & 0.9403         & \textcolor{blue}{15.36} & \textcolor{red}{7.67}       \\
\multicolumn{2}{c|}{TMNet}         & \textcolor{blue}{26.23}      & \textcolor{blue}{0.8011}      & 28.78        & 0.8640       & \textcolor{blue}{33.51}         & 0.9159         & 35.60          & 0.9380          & \textcolor{blue}{37.04}         & \textcolor{blue}{0.9435}         & 14.69 & 12.26       \\ \hline
\multicolumn{2}{c|}{STDAN (Ours)}         & \textcolor{red}{26.28}      & \textcolor{red}{0.8041}      & \textcolor{blue}{28.94}        & \textcolor{red}{0.8687}       & \textcolor{red}{33.66}         & \textcolor{red}{0.9176}         & \textcolor{red}{35.70}          & \textcolor{red}{0.9387}          & \textcolor{red}{37.10}         & \textcolor{red}{0.9437}         & 13.80 & \textcolor{blue}{8.29}       \\ \hline
\end{tabular}
\label{tab:sota}
\end{table*}

\subsection{High-Resolution Frame Reconstruction}

To reconstruct HR frames from the enhanced features $\left\{F_t^{*}\right\}_{t=1}^{2N-1}$, we first employ $m_b$ RSTBs \cite{swinir} to map feature $F_t^*$ to deep feature $F_t^d$. Then, these deep features further pass through an upsampling module to realize the HR video frames $\left\{I_{t}^{hr}\right\}_{t=1}^{2N-1}$. Specifically, the upsampling module consists of the PixelShuffle layer \cite{pixelsf} and several convolutional layers. For optimizing our proposed network, we adopt the Charbonnier function \cite{cbloss} as the reconstruction loss:
\begin{equation}\label{eq:cbloss}
    L_{rec} = \sqrt{||I_t^{hr}-I_t^{GT}||^2+\epsilon^2}\ ,
\end{equation}
where $I_t^{GT}$ indicates the ground truth of $t$-th reconstructed video frame $I_t^{hr}$, and the value of $\epsilon$ is empirically set to $1\times10^{-3}$. With the loss function, our STDAN can be end-to-end trained to generate HR slow-motion videos from corresponding LR and LFR counterparts.

\section{Experiments}

In this section, we first introduce the datasets and evaluation metrics used in our experiments. Then, the implementation details of our STDAN are elaborated. Next, we compare our proposed network with state-of-the-art methods on public datasets. Finally, we carry out ablation studies to investigate the effect of the modules adopted in our STDAN.

\begin{figure*}[t]
\centering
\includegraphics[height=12cm]{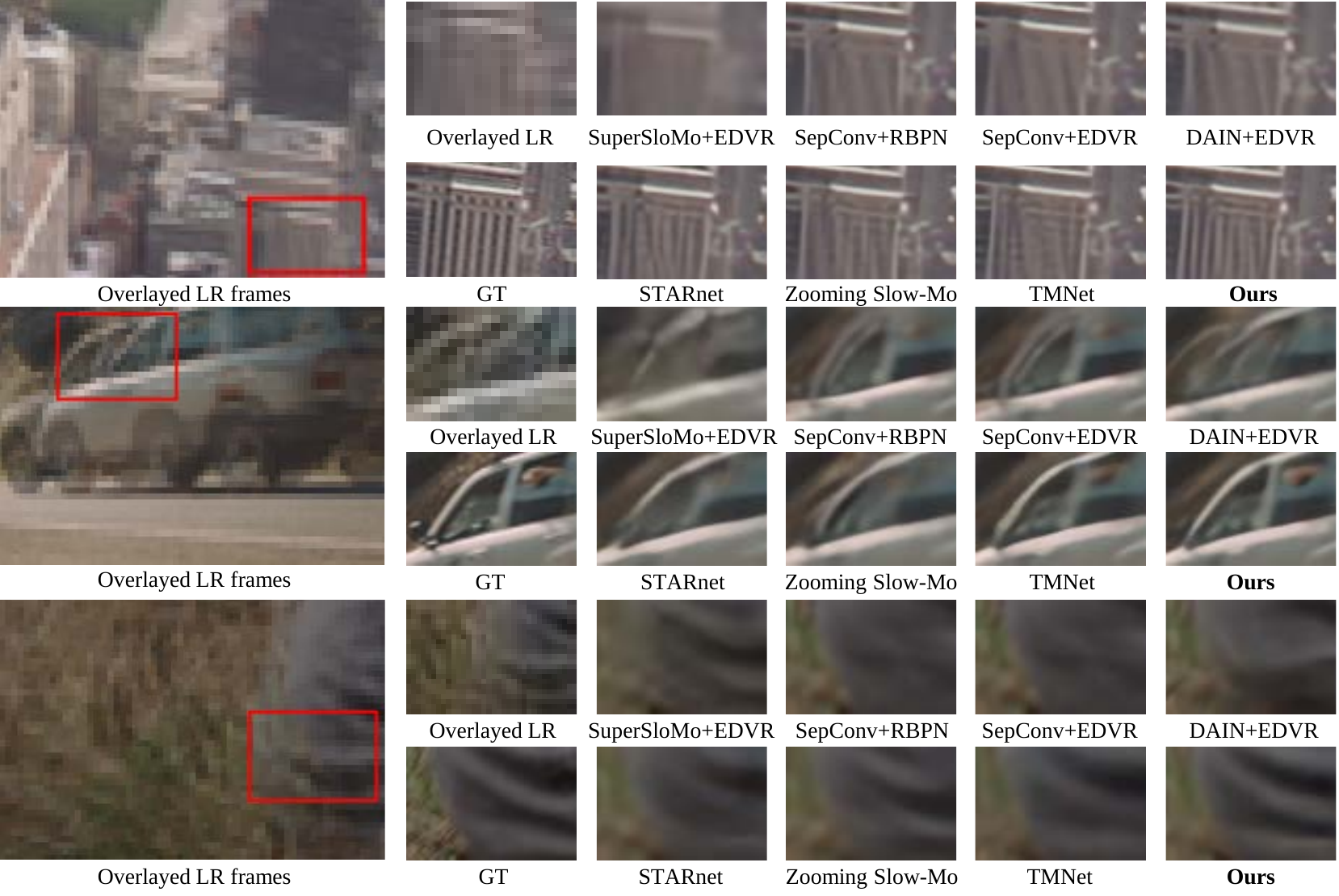}
\caption{Visual comparisons of different STVSR approaches on Vid4 and Vimeo datasets. We can see that our model can recover more accurate structures.}
\label{fig:sota_com2}
\end{figure*}


\subsection{Datasets and Evaluation Metrics}

\textbf{Datasets} We use the Vimeo-90K dataset \cite{toflow} to train our network. Specifically, the Vimeo-90K dataset consists of more than 60,000 training video sequences, and each video sequence has seven frames. We adopt the raw seven frames as our HR and HFR supervisions. The corresponding four LR and LFR frames are downscaled by a factor of 4 with bicubic sampling from these odd-numbers ones. The Vimeo-90K also provides corresponding testsets that can be divided into \emph{Vimeo-Slow}, \emph{Vimeo-Medium} and \emph{Vimeo-Fast} according to the degree of motion. The three testsets serve as the evaluation datasets in our experiments. Same as STVSR methods \cite{zooming,tmnet}, six video sequences in \emph{Vimeo-Medium} testset and three sequences in \emph{Vimeo-Slow} testset are removed to avoid infinite values on PSNR. In addition, we report the results on Vid4 \cite{vid4} and SPMC-11 \cite{drvsr} of different approaches.

\begin{figure*}[t]
\centering
\includegraphics[height=4.5cm]{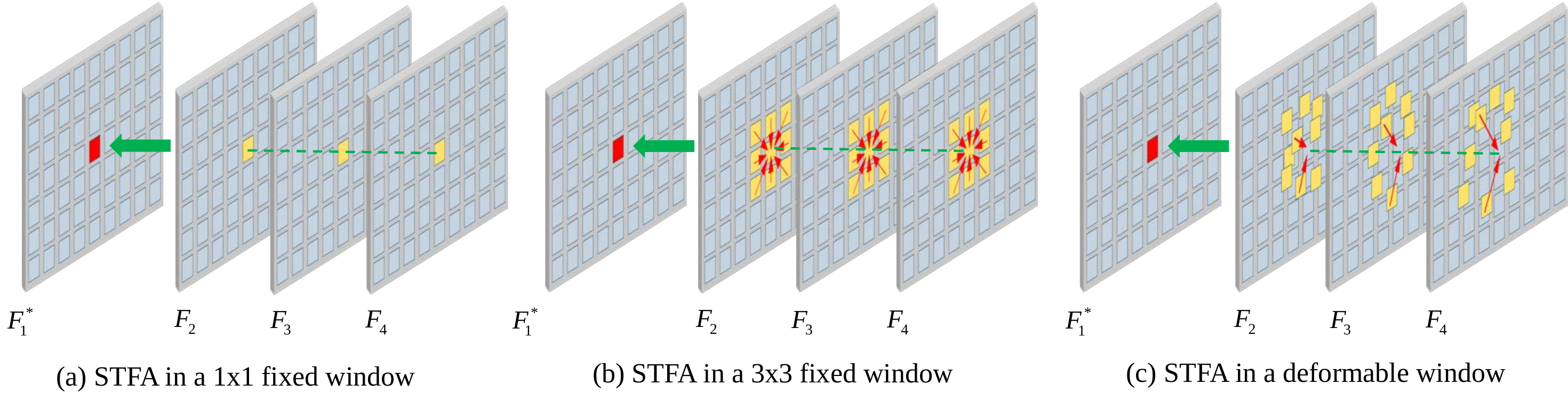}
\caption{Three different aggregation methods in the feature aggregation module. `STFA' refers to spatial-temporal feature aggregation. Note that we only show 4 frames for a illustration, and STFA in a deformable window denotes our STDFA module.}
\label{fig:demo_window}
\end{figure*}

\begin{figure*}[t]
\centering
\includegraphics[height=4.2cm]{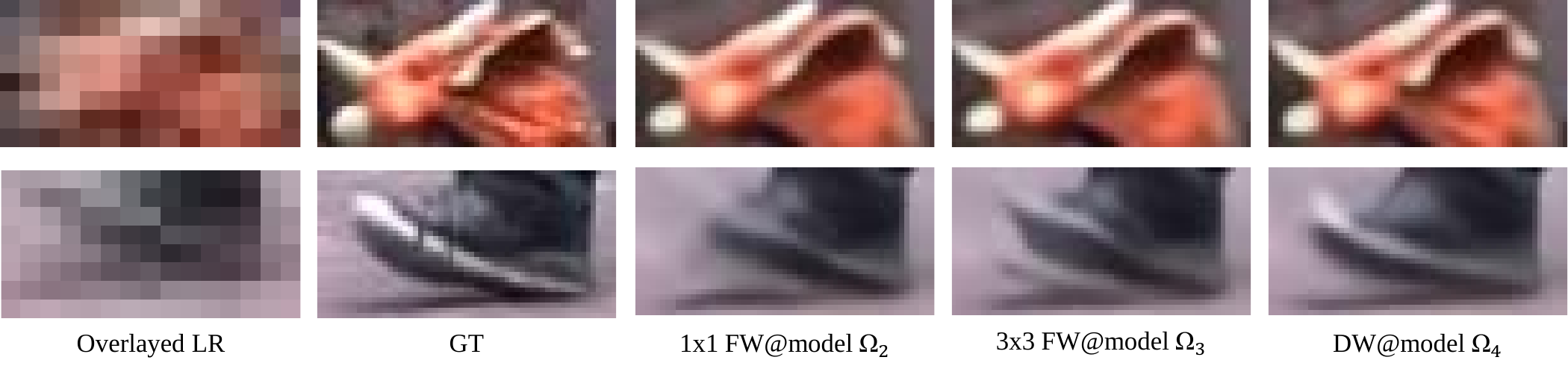}
\caption{Ablation study on the feature aggregation module. `FW' indicates fixed window, while `DW' refers to deformable window.}
\label{fig:ablation_window}
\end{figure*}

\noindent
\textbf{Evaluation Metrics} To compare diverse STVSR networks quantitatively, Peak Signal to Noise Ratio (PSNR) and Structural SIMilarity (SSIM) \cite{ssim} are adopted in our experiments as evaluation metrics. In this paper, we calculate the PSNR and SSIM metrics on the Y channel of the YCbCr color space. In addition, we also compare the parameters and inference speed of various models.

\subsection{Implementation Details}

In our STDAN, the number of RSTBs in the feature extraction module and frame feature reconstruction module is 2 and 6, respectively, where each RSTB contains 6 STBs. In addition, the number of feature and embedded feature channels are set to be 64 and 72 separately. In the LSTFI module, we utilize a Pyramid, Cascading, and Deformable (PCD) structure in \cite{edvr} to achieve DFI. The hidden states in the forward and backward branches are initialized to zeros. In the STDFA module, the value of $k$ and $T$ are set to 3 and 2, respectively. We augment the training frames by randomly flipping horizontally and $90^\circ$ rotations during the training process. Then, we crop the input LR frames with a size of $32\times32$ at random to the network, and the batch size is set to be 18. Our model is trained by Adam \cite{adam} optimizer by setting $\beta_1 = 0.9$ and $\beta_2 = 0.999$. We employ cosine annealing to decay the learning rate \cite{cosine} from $2e-4$ to $1e-7$. We implement the STDAN with PyTorch and train our model on 6 NVIDIA GTX-1080Ti GPUs.

\begin{figure*}[t]
\centering
\includegraphics[height=3.9cm]{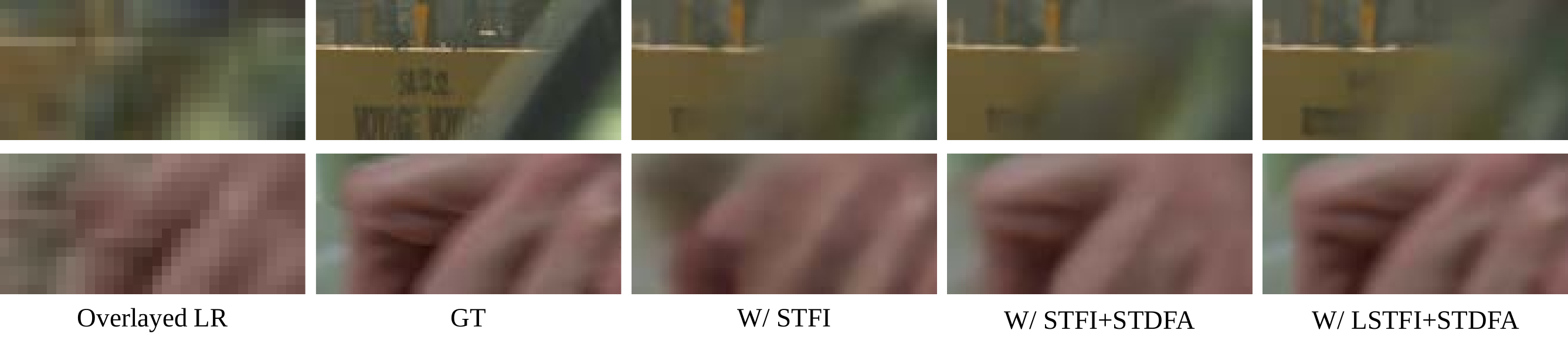}
\caption{Ablation study on the proposed modules. We can see that STDFA can effectively suppress blurring artifacts and recover correct visual structures and the LSTFI can further help to reconstruct fine details.}
\label{fig:ablation_study}
\end{figure*}

\subsection{Comparison with State-of-the-art Methods}

We compare our STDAN with existing state-of-the-art (SOTA) one-stage STVSR approaches: STARnet \cite{starnet}, Zooming Slow-Mo \cite{zooming}, RSTT \cite{geng2022rstt} and TMNet \cite{tmnet}. In addition, we also compare the performance of our network with SOTA two-stage STVSR algorithms, like Zooming Slow-Mo \cite{zooming} and TMNet \cite{tmnet}. Specifically, two-stage STVSR methods are composed of SOTA VFI and SR algorithms. These VFI networks are SuperSloMo \cite{superslomo}, SepConv \cite{sepconv} and DAIN \cite{dain}, respectively, while SOTA SR approaches are RCAN \cite{sisr-rcan}, RBPN \cite{rbpn} and EDVR \cite{edvr}. 

Quantitative results of various STVSR methods are shown in Table \ref{tab:sota}. From the table, we can see that: (1) Our STDAN with fewer parameters obtains SOTA performance on both Vid4 \cite{vid4} and Vimeo \cite{toflow}; (2) For the SPMC-11 \cite{drvsr} dataset, our model is only 0.1dB lower than STARnet \cite{starnet} in terms of PSNR, but our STDAN acquires better results than it on SSIM \cite{ssim} index, which demonstrates our network can recover more correct structures. In addition, our model only needs one thirteenth parameters of STARnet.

Visual comparison of different models are displayed in Fig. \ref{fig:sota_com2}. We observe that our STDAN, with the proposed LSTFI and STDFA modules, restores more accurate structures and fewer motion blurs compared with other STVSR approaches, which confirms the higher value on PSNR and SSIM achieved by our model.

\subsection{Ablation Study}

To investigate the effect of the proposed modules in our STDAN, we conduct comprehensive ablation studies in this section. 

\begin{table}[t]
\centering
\caption{Ablation study on the proposed modules. Our long-short term feature interpolation leverages more input LR frames to assist in the interpolation process. The proposed spatial-temporal feature aggregation in the deformable window can adaptively capture spatial-temporal contexts among different frames for HR frame reconstruction. `STFA' indicates spatial-temporal feature aggregation.}
\footnotesize
\setlength{\tabcolsep}{0.25mm}
\begin{tabular}{cc|ccccc}
\hline
\multicolumn{2}{c|}{Method}                                                                                                   & ${\rm \Omega_1}$ & ${\rm \Omega_2}$ & ${\rm \Omega_3}$ & ${\rm \Omega_4}$ & ${\rm \Omega_5}$ \\
\multicolumn{2}{c|}{Parameters (M)}                                                                                           &  5.44   &  5.54   &  5.54   &  5.82   &  8.29   \\ \hline
\multicolumn{1}{c|}{\multirow{2}{*}{\begin{tabular}[c]{@{}c@{}}Feature\\ Interpolation\end{tabular}}} & Short-term feature interpolation                   & \ding{51}    &  \ding{51}   &  \ding{51}   &  \ding{51}   &     \\
\multicolumn{1}{c|}{}                                                                                 & Long-short term feature interpolation                 &     &     &     &     & \ding{51}    \\ \hline
\multicolumn{1}{c|}{\multirow{3}{*}{\begin{tabular}[c]{@{}c@{}c@{}}Feature \\ Aggregation\end{tabular}}}  & STFA in a 1x1 fixed window      &     &  \ding{51}   &     &     &     \\
\multicolumn{1}{c|}{}                                                                                 & STFA in a 3x3 fixed window      &     &     &  \ding{51}   &     &     \\
\multicolumn{1}{c|}{}                                                                                 & STFA in a deformable window &     &     &     &  \ding{51}   & \ding{51}    \\ \hline
\multicolumn{2}{c|}{Vid4 (slow motion)}                                                                                       &  25.27   &  25.69   &  25.85   &  25.97   &  26.28   \\ \hline
\multicolumn{2}{c|}{Vimeo-Fast (fast motion)}                                                                                 &  35.88   &  36.22   & 36.41    & 36.63    & 37.10    \\ \hline
\end{tabular}
\label{tab:ab}
\end{table}

\noindent
\textbf{Feature Aggregation} To valid the effect of the proposed spatial-temporal deformable feature aggregation (STDFA) module, we establish a baseline: model ${\rm \Omega_1}$. It only adopts short-term information to perform interpolation, and then directly reconstructs HR video frames through the frame feature reconstruction module without feature aggregation process. In contrast, we compare three different models: ${\rm \Omega_2}$, ${\rm \Omega_3}$ and ${\rm \Omega_4}$ with feature aggregation. For the spatial-temporal feature aggregation process in the model ${\rm \Omega_2}$, illustrated in Fig. \ref{fig:demo_window}(a), each feature vector aggregates the information at the same position of other frame features, that is, the feature vector attends the valuable spatial content in a $1\times1$ window. We enlarge the window size of the model ${\rm \Omega_3}$ to 3. Considering large motions between frames. A deformable window is applied in the model ${\rm \Omega_4}$. As shown in Fig. \ref{fig:demo_window}(c), model ${\rm \Omega_4}$ adopts the STDFA module to perform feature aggregation. 

Quantitative results on Vid4 \cite{vid4} and \emph{Vimeo-Fast} \cite{toflow} datasets are shown in Table \ref{tab:ab}. From the table, we know that: (1) Feature aggregation module can improve the reconstruction results; (2) The larger the spatial range of feature aggregation, the more useful information can be captured to enhance recovery quality of HR frames. Qualitative results of the three models are represented in Fig. \ref{fig:ablation_window}, which confirms the feature aggregation in the deformable window can acquire more helpful content.

\noindent
\textbf{Feature Interpolation} To investigate the effect of the proposed long-short term feature interpolation (LSTFI) module, we compare two models: ${\rm \Omega_4}$ and ${\rm \Omega_5}$. As shown in Fig. \ref{fig:lstfi}, the model ${\rm \Omega_5}$ with LSTFI can exploit short-term information of two neighboring frames and long-term information of hidden states from other LSTCs. In comparison, model ${\rm \Omega_4}$ only uses two adjacent frames to interpolate the feature of the intermediate frame. From Table \ref{tab:ab}, combining long-term and short-term information can achieve better feature interpolation results, which leads to high-quality HR frames with more details, as illustrated in Fig. \ref{fig:ablation_study}.

\noindent
\textbf{Efficiency of selecting the first $T$ points} We also investigate the efficiency of determining the first $T$ points in our STDFA module. Specifically, the model's inference time of each Vimeo sequence without/with the keypoint selection are 0.542s/0.543s, which demonstrates that the utilization of the keypoint selection in our STDFA module cannot lead to a significant increase in the inference time of the model.

\begin{figure}
\centering
\includegraphics[height=3.3cm]{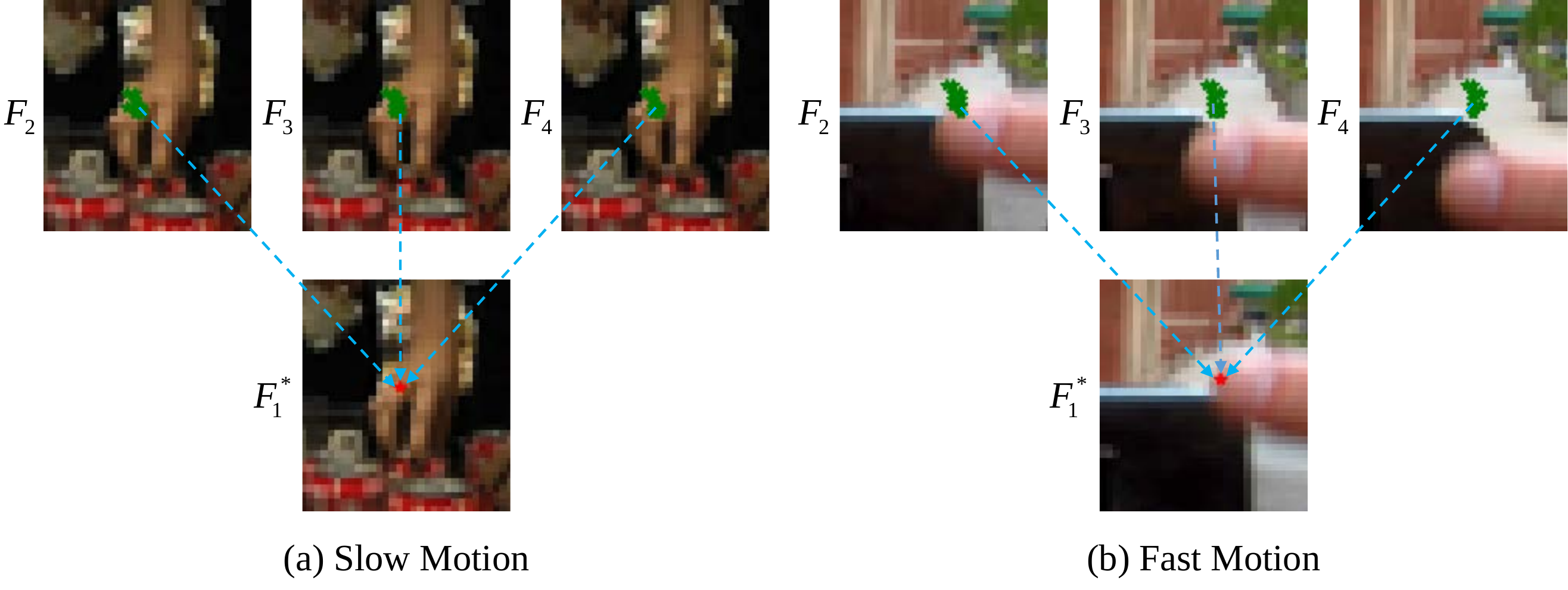}
\caption{Visualization of deformable sampling locations. The \textcolor{red}{red star} in the frame feature $F_1^*$ denotes a feature vector, and \textcolor{green}{green stars} in the other frame features indicate the corresponding sampling locations of the feature vector. Note that we directly show the sampling locations on the video frames rather than frame features, and we only show 4 frames for a better illustration.}
\label{fig:visualization}
\end{figure}

\section{Failure Analysis}

Although our method can outperform existing SOTA methods, it is not perfect especially when handling fast-motion videos. As shown in Fig. \ref{fig:visualization}, we found that our deformable attention might sample wrong locations when video motions are fast. The key reason is that the predicted deformable offsets cannot accurately capture relevant visual contexts due to the large motions.

\section{Conclusion}

In this paper, we propose a deformable attention network called STDAN for STVSR. Our STDAN can utilize more input video frames for the interpolation process. In addition, the network adopts deformable attention to dynamically capture spatial and temporal contexts among frames to enhance SR reconstruction. Thanks to the LSTFI and STDFA modules, our model demonstrates superior performance to recent SOTA STVSR approaches on public datasets.



\end{document}